\documentclass[a4paper, 10pt, conference]{ieeeconf}      %
\usepackage{FG2026}

\FGfinalcopy %

\IEEEoverridecommandlockouts                              %
\usepackage{times}
\usepackage{epsfig}
\usepackage{graphicx}
\usepackage{amsmath}
\usepackage{amssymb}

\usepackage{float}

\usepackage[table]{xcolor}
\usepackage{colortbl}
\usepackage{booktabs, makecell}
\usepackage{footmisc}
\usepackage[compress]{cite}
\usepackage[pagebackref=false,breaklinks=true,colorlinks=true,citecolor=green,bookmarks=false]{hyperref}
\usepackage{cleveref}
 \usepackage[verbose]{placeins} 
 \usepackage{multirow}
\makeatletter
\@ifundefined{labelindent}{}{\let\labelindent\relax}
\makeatother
 \usepackage{enumitem}

\usepackage{soul}

\usepackage{nicematrix}

\usepackage[accsupp]{axessibility}

\usepackage{makecell}

\usepackage{pifont}

\usepackage{arydshln}   %

\usepackage{tikz}
\usetikzlibrary{patterns}
\usetikzlibrary{patterns.meta}
\definecolor{grpA}{RGB}{166,206,227}
\definecolor{grpB}{RGB}{178,223,138}

\usepackage{caption}

\newcommand{\dashmidrule}{%
  \noalign{\vskip\aboverulesep}%
  \hdashline[2pt/5pt]%
  \noalign{\vskip\belowrulesep}%
}

\title{\LARGE \bf
FunFace: Feature Utility and Norm Estimation for Face Recognition %
}

\author{\parbox{16cm}{\centering
    {\normalsize %
    Žiga Babnik$^1$, Fadi Boutros$^2$, Naser Damer$^{2,3}$, Deepak Kumar Jain$^4$, Peter Peer$^1$ and Vitomir Štruc$^1$
    }\\
    {\small
    $^1$University of Ljubljana, Ljubljana, Slovenia\\
    $^2$Fraunhofer Institute for Computer Graphics Research IGD, Germany\\
    $^3$Department of Computer Science, TU Darmstadt, Germany \\
    $^4$Dalian University of Technology, Dalian, China\\
    }}
    \thanks{This work is supported by the ARIS Young Researcher program, the ARIS Research Programmes P2-0250 Metrology and Biometric Systems and P2-0214 Computer Vision, as well as by the German Federal Ministry of Education and Research and the Hessen State Ministry for Higher Education, Research and the Arts within their joint support of the National Research Center for Applied Cybersecurity ATHENE.
    We thank Arnes and the SLING HPC consortium, under the EuroCC 2 project, for providing the necessary computational resources.%
    }%
}
\renewcommand{\baselinestretch}{0.995}

\usepackage{fancyhdr}
\fancypagestyle{firststyle}
{
   \fancyhf{}
   
   \fancyhead[C]{
    \textcolor{red}{
       This paper has been accepted to FG 2026. If you find this research useful, please consider citing the paper.}
    }
}

\begin{document}

\ifFGfinal
\thispagestyle{empty}
\pagestyle{empty}
\else
\author{}
\pagestyle{plain}
\fi
\maketitle
\thispagestyle{firststyle}

\begin{abstract}
Face Recognition (FR) is used in a variety of application domains, from entertainment and banking to security and surveillance. Such applications rely on the FR model to be robust and perform well in a variety of settings. To achieve this, state-of-the-art FR models typically use expressive adaptive margin loss functions, which tie the feature norm to concepts related to sample quality, such as recognizability and perceptual image quality. Recently, through the development of Face Image Quality Assessment (FIQA) techniques, biometric utility has become the preferred measure of face-image quality and has been shown to be a better predictor of the usefulness of samples for face recognition compared to more human-centric aspects, such as resolution, blur, and lighting, tied to general image quality. While image quality expressed through feature norms exhibits a certain level of correlation with biometric utility, it does not fully encapsulate all aspects of utility. To address this point, we propose a new adaptive margin loss, FunFace (Face Recognition Through Utility and Norm Estimation), which incorporates biometric utility, estimated by the Certainty Ratio, into the adaptive margin, taking inspiration from AdaFace. We show that FunFace (when used to train a face recognition model) achieves competitive results to other state-of-the-art FR models on benchmarks containing high-quality samples, while surpassing them on low quality benchmarks. The code is available at {\small \url{https://github.com/LSIbabnikz/FunFace}}.

\end{abstract}

\section{Introduction}\label{sec:introduction}

Learning discriminative features is the core focus of Face Recognition (FR) techniques \cite{fr_survey}.
Through the use of new, larger training datasets \cite{ms1m, webface260m, partialfc}, and the development of advanced adaptive loss functions \cite{curricularface, magface, elasticface, adaface}, models have achieved excellent performance on a variety of benchmark datasets. Most mentioned benchmarks focus on evaluating the model's performance using images of relatively high quality \cite{lfw, cplfw, agedb}. Recently, however, more focus has been placed on benchmarks with low-image-quality, replicating difficult security and surveillance scenarios \cite{tinyface, dronesurf, briar}. %
Domain adaptation (transfer learning) can improve performance in low-quality domains \cite{petalface}, but requires additional processing and can lead to decreased performance in high-quality domains.  %
Adaptive loss functions \cite{magface, adaface, elasticface}, on the other hand, can facilitate difficult-to-recognize images directly, through %
adaptive margin functions, enabling excellent performance across all quality-domains.

\begin{figure}
    \centering
    \includegraphics[width=\linewidth]{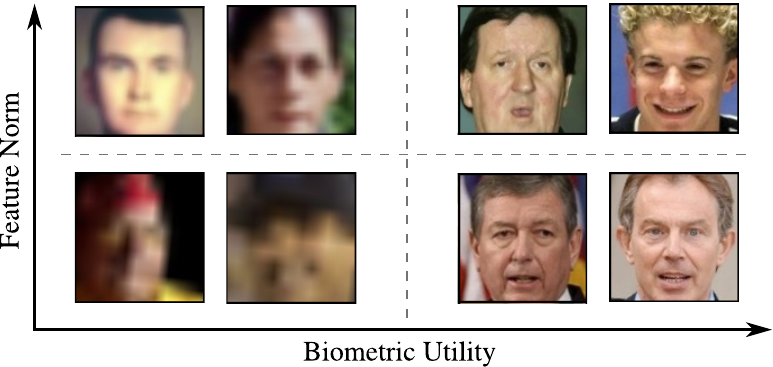}
    \caption{\textbf{Comparison of the feature norms and biometric utility of samples.} Feature norms commonly used by adaptive margin functions \cite{magface, adaface}, have been shown to correlate well with the visual image quality of samples. %
    While visual quality estimates provide a good source of information for the usefulness of samples for the recognition task, recently, biometric utility has been shown to capture the usefulness of samples more accurately.  
    Here, we show visual examples of both high and low feature norm images as estimated by AdaFace \cite{adaface}, and compare their biometric utility, estimated by CR-FIQA \cite{cr-fiqa}. We can observe a disparity between the feature norms and biometric utility, alluding to the fact that the information present in feature norms cannot accurately represent the biometric sample utility.
    \vspace{-3mm}
    }
    \label{fig:teaser}
\end{figure}

Modern FR primarily relies on margin-based loss functions, which extend the cross-entropy (softmax) loss, with a simple margin term. The two most commonly used are the angular -- ArcFace \cite{arcface} and the additive -- CosFace \cite{cosface} loss functions, both of which use a static margin value for improved separation of classes. Recently, adaptive margin functions have outperformed their static counterparts by assigning different margin values to individual samples. %
Adaptive methods generally focus on incorporating information tied to concepts such as the classifiability \cite{elasticface}, recognizability \cite{magface}, and (visual) quality \cite{adaface} of samples, into the margin function. All of these concepts are in some way related to the usefulness (or utility) of the sample for recognition. The utility, sometimes also referred to as the biometric quality \cite{eval1, eval2}, is tied to the likelihood of a sample causing a critical false match error during recognition. %
Face Image Quality Assessment (FIQA) techniques \cite{faceqnet2, sdd-fiqa, serfiq, cr-fiqa, ediffiqa, froq, faceqan} attempt to estimate the utility of samples, and show excellent performance for the task of sample quality assessment, outperforming standard visual image quality descriptors, which focus on one or more quality components, such as blur, illumination, or sharpness. %

Recent adaptive loss functions \cite{magface,adaface} incorporate additional information, tied to the difficulty of a sample, into the margin function by using %
feature norms. While feature norms have been shown to correlate well with sample visual image quality, their relationship to biometric utility remains underexplored. In Fig. \ref{fig:teaser} we showcase face samples for which the estimated feature norm and biometric utility do not agree. Here, samples with high feature norms can exhibit a large variety of utility, and vice versa. %
Motivated by the predictive power of biometric utility for sample quality estimation and the disconnect between the feature norms and utility scores, we propose a new adaptive loss function, named FunFace (\textbf{F}eature \textbf{U}tility and \textbf{N}orm Estimation for \textbf{Face} Recognition). The proposed loss function builds on the success of AdaFace \cite{adaface}, %
by incorporating an additional source of information, alongside the feature norm, into the margin function directly tied to the samples' biometric utility. This established a direct link between the samples' utility and their assigned margin value, enabling the underlying FR model to more easily adapt to samples of varying qualities, resulting in an improvement in performance, specifically in the low-quality domain.   %

\begin{figure*}[!ht]
    \centering
    \includegraphics[width=\linewidth]{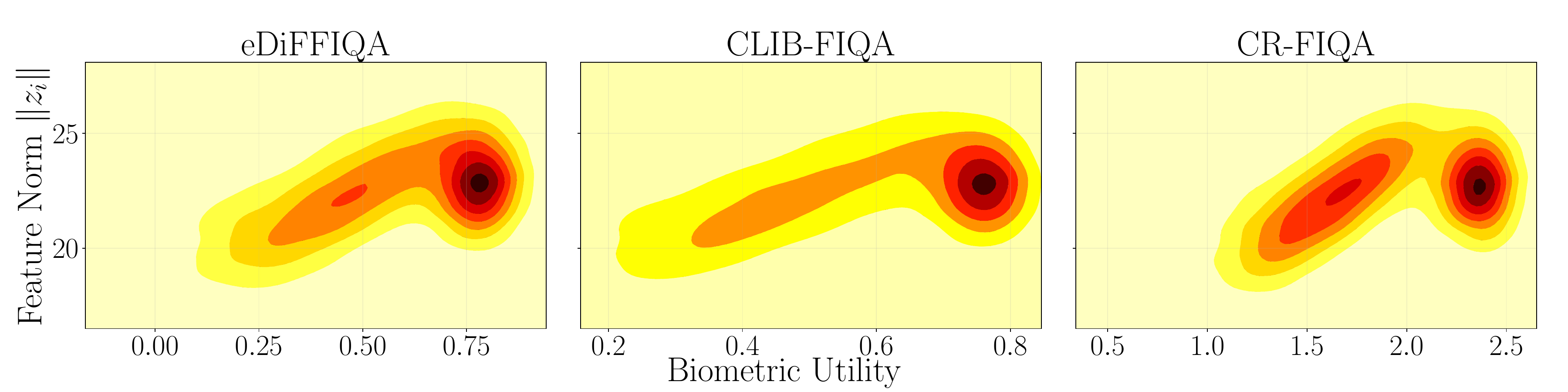}
    \caption{\textbf{Comparison between the feature norm and the utility of samples.} Using samples from the XQLFW \cite{xqlfw} benchmark, we generated density maps tying the feature norms and biometric utility of samples. We analyze the feature norms produced by AdaFace, and use three state-of-the-art FIQA techniques: eDifFIQA, CLIB-FIQA, and CR-FIQA to estimate the biometric utility of samples.  %
    We can observe a distinct shape for all included FIQA techniques, where low-scoring samples exhibit linear correlation between the feature norm and biometric utility, while higher-scoring samples exhibit an inverse correlation between the two measures. %
    \vspace{-5mm}
    }
    \label{fig:utility_v_norm}
\end{figure*}

\begin{figure}
    \centering
    \includegraphics[width=\linewidth]{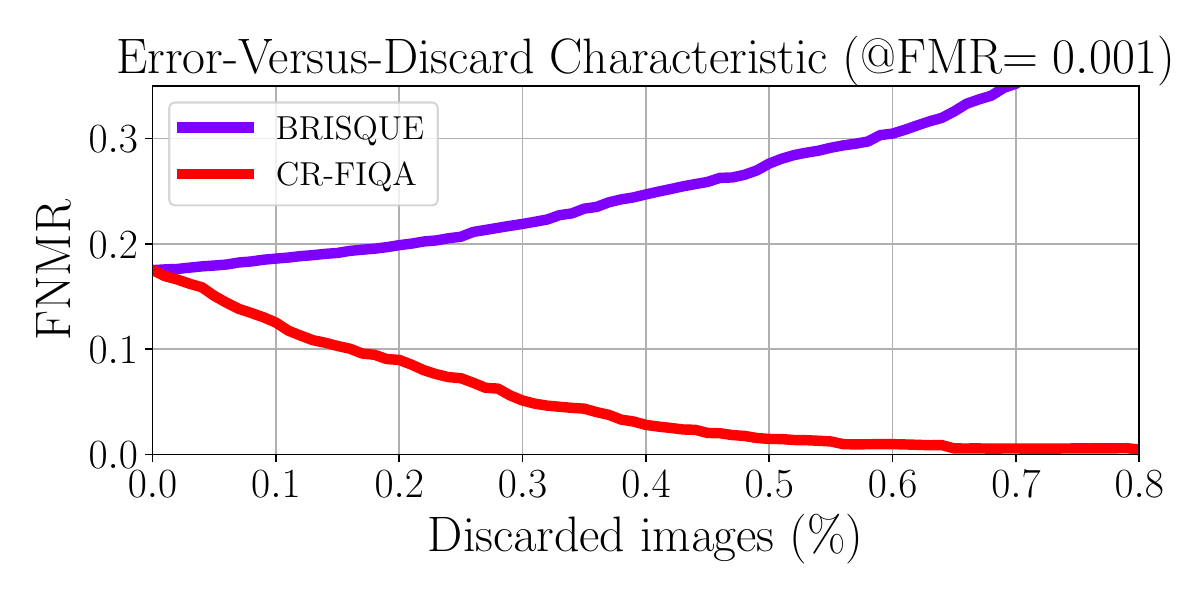}
    \caption{\textbf{Predictive strength of FIQA.} We compare BRISQUE, an IQA technique, against CR-FIQA, a supervised FIQA technique, using the Error-versus-Discard Characteristic (EDC) curves, which measure the False Non-Match Rate (FNMR) at a given False Match Rate (FMR) at various discard rates, on the XQLFW benchmark \cite{xqlfw}. %
    }
    \label{fig:iqa_v_fiqa}
\end{figure}

\section{Related Work}\label{sec:related_work}
                  
Face recognition has seen significant performance improvements, especially in low-quality domains, with the development of new deep-learning methods. While aspects such as the use of large datasets \cite{ms1m, webface260m, partialfc}, and improvements in the architectures \cite{resnet, transformer, swintransformer} can be attributed to these improvements, we focus primarily on the developments of more expressive loss functions \cite{facenet, arcface, adaface}.%

\subsection{Margin Based Loss Functions.}\label{sec:related_work:subsec:margin_loss_functions}

Modern FR models most often employ some form of cross-entropy (softmax) loss, which can be written as:
\begin{equation}\label{eq:cross_entropy}
    \mathcal{L}_{CE} = - log\left(\frac{e^{(s\cdot cos(\theta_{i}))}}{\sum_{j=1}^C e^{(s\cdot cos(\theta_{j}))}}\right),
\end{equation}
where $\theta_k$, represents the angle between the $k$-th class center and the given sample's feature, $C$ is the number of classes in the training set, $i$ is the index of the true (positive) class center, and $s$ is a scaling hyperparameter of the model.

By applying a margin to the positive class term - $cos(\theta_i)$ (PCT), we can obtain commonly used FR loss functions. Using a generalized form of $PCT$, written as:
\begin{equation}\label{eq:pct}
    PCT  = cos(m_{Sph}\cdot\theta_{i} + m_{Arc}) - m_{Cos},
\end{equation}
which can represent the three most common static margin-based loss functions currently in use. By applying only one of the included margins, we can derive the SphereFace \cite{sphereface}, ArcFace (angular) \cite{arcface}, and CosFace (additive) \cite{cosface} loss functions, using the $m_{Sph}$, $m_{Arc}$, and $m_{Cos}$ margins, respectively. %
Altering the $PCT$, we rewrite the CE loss formulation, presented in Eq. (\ref{eq:cross_entropy}), as:
\begin{equation}\label{eq:cross_entropy_margin}
    \mathcal{L}_{m} = - log\left(\frac{e^{(s\cdot PCT)}}{e^{(s\cdot PCT)} + \sum_{j=1,j\neq i}^C e^{(s\cdot cos(\theta_{j}))}}\right).
\end{equation}
Here, the terms evaluating the angle to the negative classes (contained in the sum) remain the same, while the $PCT$ is altered with the appropriate margin values.%

\subsection{Adaptive Loss Functions}\label{sec:related_work:subsec:adaptive_loss_functions}

The use of adaptive margin functions can yield significant performance gains in unconstrained recognition scenarios, where no quality guarantees can be made regarding the input samples. %
This can be attributed to the ability of such functions to assign different margin values to each training sample, where the assigned value is often correlated with the sample's difficulty in the recognition process.
CurricularFace \cite{curricularface} changes the margin between different stages (curriculum) of training, allowing it to focus on easier samples in the early stages and harder samples in later stages of training. MagFace \cite{magface} and AdaFace \cite{adaface} find that the feature norm aligns well with the recognizability and visual image quality of samples, respectively, and use it for the adaptive margin. ElasticFace \cite{elasticface}, on the other hand, focuses on adapting the margin according to the distance to the positive class center.  %

Our proposed loss function provides an additional source of information to the adaptive function, alongside the feature norm, in the form of the Certainty Ratio (CR) \cite{cr-fiqa}. %
The CR correlates well with the biometric utility of samples, and is an excellent descriptor of sample quality. Incorporating CR and feature norms allows the model to focus on visual image quality queues and the biometric quality of samples, further improving performance in low-quality domains. %

\begin{figure*}[!ht]
    \centering
    \includegraphics[width=\linewidth]{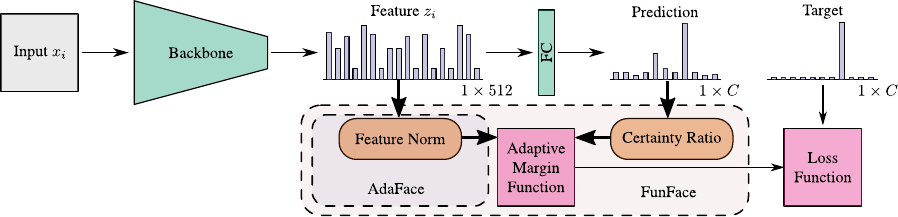}
    \caption{\textbf{Overview of the proposed FunFace loss function.} We extend the AdaFace framework, which adapts the margin according to the measured feature norm, with additional information extracted from the class similarities in the form of the Certainty Ratio. Both the feature norm and certainty ratio are combined to form the final margin value. This allows the final loss to distinguish between samples with varying feature norms as AdaFace, while providing better angular separation, which more closely follows the biometric utility of individual samples.
    \vspace{-3mm}
    }
    \label{fig:overview}
\end{figure*}

\section{Preliminaries}\label{sec:preliminaries}

\subsection{AdaFace Loss Function} \label{sec:preliminaries:subsec:adaface_loss_function}

AdaFace shows that the feature norm correlates well with the image quality of samples, as estimated by BRISQUE \cite{brisque}, a popular blind image quality estimation model. It adapts the commonly used $PCT$, presented in Eq. (\ref{eq:pct}), using:
\begin{equation}\label{eq:adaface}
    PCT_{Ada} = cos\left(\theta_{i} + g_{angle}\right) - g_{add},
\end{equation}
where $g_{angle}$ and $g_{add}$ are the adaptive angular and additive margin functions. The feature norm is integrated into these margin functions, as shown here: 
\begin{equation}\label{eq:adaface_margin}
    g_{angle} = -m\cdot\hat{{\|z_i\|}},\;\; g_{add} = m + (m \cdot \hat{\|z_i\|}),
\end{equation}
where $\hat{\|z_i\|}$ is the normalized feature norm, and $m$ is a static margin value. By defining the angular and additive margin functions in such a way, AdaFace can be seen as a generalized version of both ArcFace and CosFace. When the normalized feature norm $\hat{\|z_i\|}$ equals $0$, the angular margin function $g_{angle}$ evaluates to $0$, and we are left only with a static additive margin, as in the CosFace loss function. Similarly, when the normalized feature norm $\hat{\|z_i\|}$ equals $-1$, the additive margin function $g_{add}$ evaluates to $0$, leaving us with only a static angular margin, as in the ArcFace loss function.
The normalized feature norm is calculated using:
\begin{equation}\label{eq:logit_norm}
    \hat{\|z_i\|} = \left\lfloor\frac{\|z_i\| - \mu_{z}}{\sigma_{z}/h} \right\rceil_{-1}^{1},
\end{equation}
where $\mu_z$ and $\sigma_z$ are the running mean and standard deviation of the feature norms, and $h$ is a hyperparameter defining the spread of the normalized distribution. The running mean and standard deviation are tracked using a simple EMA (Exponential Moving Average) approach. %

\subsection{CR-FIQA} \label{sec:preliminaries:subsec:cr-fiqa}

Estimating the biometric utility of samples has proven to be extremely useful for recognition in low-quality domains, where discarding and quality-weighted averaging of samples can be utilized. %
CR-FIQA \cite{cr-fiqa} is a supervised FIQA technique that does not require any pseudo-quality labels. During the training of a FR model, the method computes the Certainty Ratio ($CR$), which is closely tied to the biometric utility, as demonstrated by its excellent performance as an estimator of sample quality. %
During training of a FR model, $CR$ can be computed as:
\begin{equation}\label{eq:cr}
    CR = \frac{CCS}{NNCCS + (1 + \epsilon)},
\end{equation}
where $CCS$ is the Class Center Angular Similarity, and $NNCCS$ is the Nearest Negative Class Center Angular Similarity. The ratio is expanded with the $(1 + \epsilon)$ term to ensure stable computation, pushing the value range from $[-1, 1]$ to $[\epsilon, 2+\epsilon]$. Expressing $CR$ directly with cosine similarities, we can write:%
\begin{equation}\label{eq:cr_ada}
    CR = \frac{cos(\theta_i)}{max_{j,j\neq i}^C(\theta_j) + (1 + \epsilon)},
\end{equation}
where $i$ is the positive class index, and $C$ is the total number of classes in the training dataset.

\begin{figure*}
    \centering
    \includegraphics[width=0.9\linewidth]{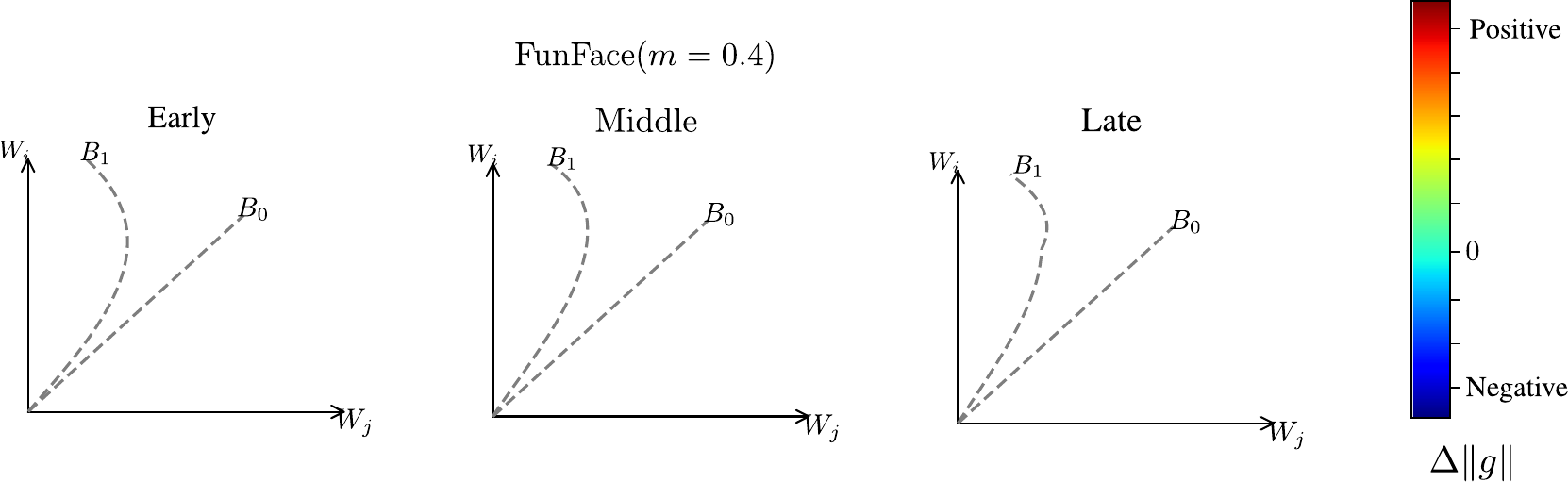}
    \caption{\textbf{Analysis of the gradient scale difference between FunFace and the AdaFace loss function.} We present the difference of the gradient scale between the AdaFace and the proposed FunFace loss function highlighing key differences between the two. Since the mean of the normalized certainty ratio changes drastically during training, we include an analysis from early, middle and late stages of training. Each figure shows the position of the positive class center $W_i$ and the nearest negative class center $W_j$, as well as the class border before applying the margin function $B_0$ and after application $B_1$. %
    \vspace{-3mm}
    }
    \label{fig:margin_feature_space}
\end{figure*}

\section{Methodology}\label{sec:methodology}

The performance of FR methods has seen a steady improvement in recent years, thanks to the development of novel adaptive loss functions \cite{curricularface, magface, adaface, elasticface}. These loss functions enable the margin to adapt to individual training samples, usually through information present in the feature norm, which can be associated with the visual image quality of samples \cite{adaface}. %
Recently, through the developments of new FIQA techniques, biometric utility has been shown to be an excellent predictor of sample quality for recognition tasks \cite{eval2, survey}, outperforming classical (visual) image quality estimation techniques \cite{niqe, brisque, rankiqa}. %
In this section, we present our proposed FunFace loss function, which integrates information about the utility of a sample, estimated by a FIQA technique %
into the adaptive margin. To further motivate the use of FIQA techniques for adaptive functions, we first demonstrate the performance gap between IQA and FIQA techniques for sample quality estimation, and show that the feature norm is an inadequate source of information for accurate estimation of a sample's utility.

\subsection{Comparison of Quality Estimation Techniques}\label{sec:methodology:subsec:iq_v_fiq}

Estimating the quality of samples is an important task for end applications dealing with variable or low-quality images. By using quality estimates, FR systems can discard samples below a specific threshold, which are more likely to cause problematic false match errors, and in turn improve their reliability. In such cases, the predictive strength of the quality estimates is an extremely important factor as it determines the potential improvement in robustness of the FR system. %

Recently, FIQA techniques have shown to be excellent predictors of sample quality for recognition, outperforming standard (visual) image quality estimators. Unlike standard approaches, which focus on aspects closely related to human vision, such as the resolution or illumination, these methods estimate the biometric utility (usefulness) of samples, concerned more so with how the FR model perceives the samples. To highlight the effectiveness of FIQA techniques and consequently of biometric utility for FR, we evaluate BRISQUE \cite{brisque} and CR-FIQA \cite{cr-fiqa} using standard Error-versus-Discard Characteristic (EDC) curves. These curves show the predictive strength of the quality estimates by iteratively discarding a percentage of the lowest quality images, and evaluating the performance of the underlying FR model at each iteration. We chose BRISQUE as the general IQA technique, since it is used in AdaFace as a predictor of image quality. The results of the analysis on the XQLFW benchmark using the AdaFace model are presented in Fig. \ref{fig:iqa_v_fiqa}, and show a clear divide between the methods. The performance of the underlying FR model gradually improves when discarding samples through CR-FIQA scores, while for BRISQUE, the performance actually drops, as seen by the increase in the False Non Match Rate (FNMR). This difference in performance between the methods demonstrates the predictive strength of FIQA techniques for sample quality estimation and further motivates the use of such techniques to guide the model during training. %

\subsection{Analysis of the Feature Norm}\label{sec:methodology:subsec:feature_norm_analysis}

The feature norm commonly used by adaptive loss functions as a source of information tied to the samples' difficulty is correlated with the visual image quality, as showcased in \cite{adaface}. However, as shown in Section \ref{sec:methodology:subsec:iq_v_fiq}, visual quality is not as important as biometric utility for sample quality estimation, leading us to explore possible correlations between the feature norm and biometric utility of samples. 

In Fig. \ref{fig:utility_v_norm} we present a comparison between the feature norms estimated by AdaFace and the biometric utility, estimated by three unique state-of-the-art FIQA techniques eDifFIQA \cite{ediffiqa}, CLIB-FIQA \cite{clibfiqa}, and CR-FIQA \cite{cr-fiqa} using images from the XQLFW benchmark. %
From the presented density maps, %
we can observe a similar pattern across all three different FIQA techniques. Because of the unique shape of the observed pattern, we can conclude that the feature norm alone does not contain enough information to accurately depict the biometric utility of a sample. For example, a sample with a high feature norm could correspond to a wide degree of different biometric utilities.
Therefore, a more explicit source of biometric utility information is required in the adaptive margin, to account for the observed discrepancy between the two measures.%

\subsection{FUNFace Loss}\label{sec:methodology:subsec:our_loss}

The proposed FunFace loss function extends on the AdaFace framework, presented in Section \ref{sec:preliminaries:subsec:adaface_loss_function}, by incorporating a biometric utility estimator directly into the adaptive margin function, as shown in Fig. \ref{fig:overview}. %
Due to its excellent performance and ease of integration, we chose CR-FIQA, described in Section \ref{sec:preliminaries}, as the target biometric utility estimator. The FunFace loss, similarly to AdaFace, defines the adaptive margin functions as: 
\begin{equation}\label{eq:our_margin_functions}
    g_{angle} = -m \cdot \kappa,\;\;\; g_{add} = m + (m \cdot \kappa),
\end{equation}
where $\kappa$ is a function of the normalized feature norm $\hat{\|z_i\|}$, as in AdaFace but also of the normalized certainty ratio $\hat{\|CR_i\|}$. The adaptive function $\kappa$ can be written as: %
\begin{equation}\label{eq:our_kappa}
    \kappa = \lambda \cdot \hat{\|z_i\|} + (1 - \lambda) \cdot \hat{\|CR_i\|},
\end{equation}
where $\lambda$ is a hyperparameter of the model called the mixing factor, which determines the scaling between the two quality descriptors $\hat{\|z\|}$ and $\hat{\|CR_i\|}$. %
To compute the $CR$ descriptor of the training samples, we adapt the original formulation, presented in Eq. (\ref{eq:cr}), as shown here:
\begin{equation}\label{eq:our_cr}
    CR_i = \frac{\lfloor CSS \rceil_{0}^{1}}{\lfloor NNCCS \rceil_{0}^{1} + \epsilon} = \frac{\lfloor cos(\theta_i) \rceil_{0}^{1} }{\lfloor max_{j,j\neq i}^C(cos\theta_j) \rceil_{0}^{1} +\epsilon},
\end{equation}
altering the normalization using $\epsilon$, and limiting all cosine values to the interval $[0, 1]$. Through testing, we find that limiting the values impacts training only very early on, and has little effect otherwise, since both the positive and the closest negative class are close enough to the sample to fall into the range of $[0, 1]$. Similar to the feature norm, presented in Eq. (\ref{eq:logit_norm}), we normalize the $CR$ values as follows: %
\begin{equation}\label{eq:our_norm_cr}
    \hat{\|CR_i\|} = \left\lfloor\frac{\|CR_i\| - \mu_{CR}}{\sigma_{CR}/h} \right\rceil_{-1}^{1},
\end{equation}
where the $\mu_{CR}$ is the running mean and $\sigma_{CR}$ the standard deviation of the samples certainty ratio, and $h$ a hyperparameter controlling the spread of the distribution. %

\subsection{Analysis of the Gradient Scale}\label{sec:methodology:subsec:our_loss}

The proposed loss focuses on visual image quality and biometric utility through the use of the feature norm and certainty ratio, respectively. The effects of the feature norm on the gradient scale using the AdaFace loss function have been studied extensively in \cite{adaface}. Since our proposed loss function directly extends AdaFace, the effects of the feature norm term should correlate well with the original formulation. However, we also include a normalized certainty ratio term in our adaptive margin function. 

The certainty ratio is fully dependent on the angular position; the norm of the feature does not affect it, but rather only the angular distance to the two adjacent class centers. In Fig. \ref{fig:margin_feature_space} we present an analysis of the impact of the certainty ratio on the gradient scale. %
The figure presents the difference in the gradient scale between the FunFace and AdaFace loss functions, showcasing how the addition of the certainty ratio influences the final decision space. Due to the dynamic nature of the normalized certainty ratio, the differences between the two change during training. Earlier on, the mean of the certainty ratio is close to one, i.e., the features are randomly spaced between the positive and nearest negative class center; however, as the training progresses, the ratio rises, leading training samples away from the negative and closer to the positive class center. From the results, we can see that earlier on FunFace puts less emphasis on more difficult samples, as highlighted by the region with negative gradient scale change, shown in blue, between the two decision boundaries. Gradually, through the training process, more weight is put on difficult samples (far from the boundary $B_1$), as highlighted by an increase in the gradient scale difference (green), for the middle and late stages of training. An inverse effect can be seen for samples on the positive side of the decision boundary $B_1$. Earlier on, FunFace emphasizes easier samples (red), while later on, this effect dissipates significantly. %

\begin{table}[!t!]
    \centering
    \renewcommand{\arraystretch}{1.1}
    \caption{\textbf{Comparison of benchmarks datasets.} We present additional information about the benchmark datasets, showcasing the differences in scale and difficulty. %
    \vspace{1mm}}%
\resizebox{\columnwidth}{!}{%
    \begin{tabular}{l r r r r r l }
\toprule

        \multirow{ 2}{*}{\textbf{Dataset}} & \multirow{ 2}{*}{\textbf{Images}} & \multirow{ 2}{*}{\textbf{IDs}} & \multicolumn{2}{c}{\textbf{Comparisons}} && \multirow{2}{*}{\textbf{Difficulty}}\\%
        \cline{4-5}%
        
        & & & Mated & Non-mated &&  \\ 
        \midrule
        LFW~\cite{lfw} & $13{\small,}233$ & $5{\small,}749$ & $3{\small,}000$ & $3{\small,}000$ && Low  \\ 
        CFP-FP~\cite{cfp-fp} & $7{\small,}000$ & $500$ & $3{\small,}500$ & $3{\small,}500$ && Low    \\
        CPLFW~\cite{cplfw} & $11{\small,}652$ & $3{\small,}930$& $3{\small,}000$ & $3{\small,}000$ && Low  \\
        AgeDB~\cite{agedb} & $16{\small,}488$ & $570$ & $10{\small,}000$ & $10{\small,}000$ && Low  \\
        CALFW~\cite{calfw} & $12{\small,}174$ & $4{\small,}025$ & $3{\small,}000$ & $3{\small,}000$ &&Low \\

        \dashmidrule  
        IJB-C~\cite{ijbc} & $138{\small,}836$ & $3{\small,}531$ & $19{\small,}557$ & $15{\small,}638{\small,}932$ && Medium \\

        \dashmidrule  
        
        TinyFace~\cite{tinyface} & $169{\small,}403$ & $5{\small,}139$ & $3{\small,}728$ & - && High \\
        DroneSURF~\cite{dronesurf} & $411{\small,}451$ & $58$ & $712{\small,}198$ & - && High \\
        SurvFace~\cite{survface} & $463{\small,}507$ & $15{\small,}573$ & $5{\small,}320$ & -  && High\\

        \bottomrule
        \vspace{-6mm}
    \end{tabular}
    }
    \label{tab:dataset_info}
\end{table}

\begin{table*}[!t]
    \centering
        \caption{\textbf{Experimental results on low and medium difficulty benchmarks.} For the low difficulty bencmarks the results are presented using the verification accuracy ($\%$), and for the mixed difficulty using the True Acceptance Rate (TAR) at three False Acceptance Rates (FARs): $1e^{-4}$, $1e^{-5}$, $1e^{-6}$. For each training dataset, we mark the \textbf{best}, and \textit{second-best} result.  %
    }
      \renewcommand{\arraystretch}{1.05}
    \resizebox{\linewidth}{!}{%
\begin{NiceTabular}{l | c | c | c | c | c | c || c | c | c }
\toprule
 \multirow{2}{*}{\textbf{Method}} & \multirow{2}{*}{\textbf{Dataset}} & \multicolumn{1}{c}{\multirow{2}{*}{\textbf{LFW}}} & \multicolumn{1}{c}{\multirow{2}{*}{\textbf{CFP-FP}}} & \multicolumn{1}{c}{\multirow{2}{*}{\textbf{CPLFW}}} & \multicolumn{1}{c}{\multirow{2}{*}{\textbf{AgeDB}}} & \multicolumn{1}{c}{\multirow{2}{*}{\textbf{CALFW}}} & \multicolumn{3}{c}{{\textbf{IJB-C (TAR@FAR=?)}}} \\
& & & & & & & $1e^{-4}$ & $1e^{-5}$ & $1e^{-6}$ \\
\midrule

CosFace(m=0.35)\cite{cosface} & \multirow{12}{*}{MS1MV2} & $99.81$ & $98.12$ & $92.28$ & $98.11$ & $95.76$ & $96.37$ & - & -\\
ArcFace(m=0.50)\cite{arcface} &  & $99.83$ & $98.27$ & $92.08$ & $98.28$ & $95.45$ & $96.03$ & -  & - \\
MV-SoftMax\cite{mvsoftmax} &  & $99.80$ & $98.28$ & $92.83$ & $97.95$ & $96.10$ & $95.20$ & - & - \\
CurricularFace\cite{curricularface} &  & $99.80$ & $98.37$ & $93.13$ & $98.32$ & $96.20$ & $96.10$ & - & - \\
BroadFace\cite{broadface} &  & $\mathbf{99.85}$ & $98.63$ & $93.17$ & $\mathbf{98.38}$ & $\mathit{96.20}$ & $96.03$ & $\mathit{94.11}$ & $\mathbf{90.41}$ \\
MagFace\cite{magface} &  & $99.83$ & $98.46$ & $92.87$ & $98.17$ & $96.15$ & $95.97$ & $94.08$ & $\mathit{90.24}$ \\
ElasticFace-Arc+\cite{elasticface} &  & $99.82$ & $98.60$ & $93.28$ & $\mathit{98.35}$ & $96.17$ & $96.40$ & - & - \\
ElasticFace-Cos+\cite{elasticface} &  & $99.80$ & $\mathbf{98.73}$ & $93.23$ & $98.28$ & $96.18$ & $96.65$ & - & -\\
PartialFC(ArcFace)\cite{partialfc} &  & $\mathit{99.83}$ & $98.45$ & $93.00$ & $98.20$ & $96.18$ & $96.20$ & - & - \\
PartialFC(CosFace)\cite{partialfc} &  & $99.83$ & $98.51$ & $93.10$ & $98.03$ & $96.20$ & $96.40$ & - & - \\
AdaFace(m=0.4)\cite{adaface} &  & $99.82$ & $98.49$ & $\mathit{93.53}$ & $98.05$ & $96.08$ & $\mathbf{96.89}$ & - & - \\ 

\rowcolor{lightgray!30}\textbf{FunFace (ours)} &  &  $99.82$& $\mathit{98.67}$& $\mathbf{93.83}$& $98.13$& $\mathbf{96.25}$ & $\mathit{96.82}$ & $\mathbf{94.47}$ & $89.01$\\

\cmidrule(lr){1-10}

VPL-ArcFace\cite{vplarcface} & \multirow{3}{*}{MS1MV3} & $\mathit{99.83}$ & $\mathit{99.11}$ & $93.45$ & $\mathbf{98.60}$ & $\mathit{96.12}$ & $96.76$ & - & -\\
AdaFace\cite{adaface} &  & $99.83$ & $99.03$ & $\mathit{93.93}$ & $98.17$ & $96.02$ & $\mathit{97.09}$ & - & -\\
\rowcolor{lightgray!30}\textbf{FunFace (ours)} &  &  $\mathbf{99.83}$& $\mathbf{99.16}$& $\mathbf{94.12}$& $\mathit{98.27}$& $\mathbf{96.15}$ & $\mathbf{97.20}$ & $\mathbf{95.54}$ & $\mathbf{90.03}$\\
\cmidrule(lr){1-10}

ArcFace\cite{arcface} & \multirow{3}{*}{WebFace4M} & $\mathit{99.83}$ & $\mathbf{99.19}$ & $94.35$ & $\mathbf{97.95}$& $96.00$ & $97.16$ & - & - \\
AdaFace\cite{adaface} &  & $99.80$ & $\mathit{99.17}$ & $\mathit{94.63}$ & $97.90$ & $\mathit{96.05}$ & $\mathbf{97.39}$ & - & -\\

\rowcolor{lightgray!30}\textbf{FunFace (ours)} &  & $\mathbf{99.85}$& $99.14$& $\mathbf{94.63}$& $\mathit{97.90}$& $\mathbf{96.15}$ & $\mathit{97.31}$ & $\mathbf{95.62}$ & $\mathbf{90.81}$\\

\bottomrule  

\end{NiceTabular}
}
\label{tab:sota_small_and_mixed_scale}
\end{table*}

\begin{table}[!t]
    \centering
    \caption{\textbf{Experimental results on the TinyFace benchmark.} %
   We present the Rank-1 and Rank-5 scores for each training dataset; we mark the \textbf{best} and \textit{second-best} result.
    }
      \renewcommand{\arraystretch}{0.5}
    \resizebox{\linewidth}{!}{%
\begin{NiceTabular}{l | c | c | c }

\toprule

\textbf{Method} & \textbf{Dataset} & \multicolumn{1}{c}{\textbf{Rank-1}} & \multicolumn{1}{c}{\textbf{Rank-5}} \\
 \midrule

URL\cite{url} & \multirow{4}{*}{MS1MV2} & $63.89$ & $68.67$ \\
CurricularFace\cite{curricularface} &  & $63.68$ & $67.65$  \\
AdaFace(m=0.4)\cite{adaface} &  & $\mathit{68.21}$ & $\mathit{71.54}$  \\
\rowcolor{lightgray!30}\textbf{FunFace (ours)} &  &  $\mathbf{69.07}$&$\mathbf{72.21}$ \\ 

\cmidrule(lr){1-4} 

AdaFace(m=0.4)\cite{adaface} &  & $\mathit{67.81}$ & $\mathit{70.98}$  \\
\rowcolor{lightgray!30}\textbf{FunFace (ours)} & \multirow{-2}{*}{MS1MV3} &  $\mathbf{69.88}$&$\mathbf{73.04}$ \\ 

\cmidrule(lr){1-4} 

ArcFace\cite{arcface} & \multirow{3}{*}{WebFace4M} & $71.11$ & $74.38$ \\
AdaFace(m=0.4)\cite{adaface} &  & $\mathit{72.02}$ & $\mathit{74.52}$ \\
\rowcolor{lightgray!30}\textbf{FunFace (ours)} &  &  $\mathbf{73.31}$& $\mathbf{76.05}$ \\

\bottomrule

\end{NiceTabular}
}
\label{tab:sota_low_quality}
\end{table}

\begin{table*}[!t]
    \centering
    \caption{\textbf{Experimental results on the high difficulty DroneSURF and SurvFace benchmarks.} %
    We present the Rank-N($\%$) score for DroneSURF and the True Acceptance Rate (TAR) at $30\%$, $20\%$, $10\%$, $1\%$ False Acceptance Rates (FARs).%
    }
      \renewcommand{\arraystretch}{0.8}
    \resizebox{0.93\linewidth}{!}{%
\begin{NiceTabular}{l | c || c | c || c | c || c | c | c | c}

\toprule

 \multirow{2}{*}{\textbf{Method}} & \multirow{2}{*}{\textbf{Dataset}} & \multicolumn{2}{c}{\textbf{DroneSURF-A}} & \multicolumn{2}{c}{\textbf{DroneSURF-P}} & \multicolumn{4}{c}{\textbf{SurvFace (TAR@FAR=?)}} \\
 & & \multicolumn{1}{c}{Rank-1} & \multicolumn{1}{c}{Rank-5} & \multicolumn{1}{c}{Rank-1} & \multicolumn{1}{c}{Rank-5} & \multicolumn{1}{c}{$30\%$} & \multicolumn{1}{c}{$20\%$} & \multicolumn{1}{c}{$10\%$} & \multicolumn{1}{c}{$1\%$} \\
\midrule

MagFace\cite{magface} & \multirow{4}{*}{MS1MV2} & $56.92$ & $68.57$  & $25.07$ & $37.27$ & $\mathit{58.15}$ & $\mathit{47.04}$ & $\mathit{32.53}$ & $9.45$ \\
ElasticFace\cite{elasticface} &  & $58.23$ & $69.94$  & $26.42$ & $38.70$ & $56.76$ & $46.18$ & $32.38$ & $\mathit{10.37}$  \\
AdaFace\cite{adaface} &  & $\mathit{59.00}$ & $\mathit{70.37}$  & $\mathit{27.51}$ & $\mathit{39.59}$ & $55.97$ & $45.39$ & $29.71$ & $7.55$ \\
\rowcolor{lightgray!30}\textbf{FunFace (ours)} &   &  $\mathbf{65.20}$ & $\mathbf{75.76}$  & $\mathbf{32.65}$ & $\mathbf{44.46}$ &$\mathbf{60.50}$ & $\mathbf{48.83}$ & $\mathbf{34.39}$ & $\mathbf{11.91}$ \\ 

\cmidrule(lr){1-10}

Adaface\cite{adaface} &  &  $\mathit{63.03}$ & $\mathit{73.86}$  & $\mathit{31.30}$ & $\mathit{43.64}$ & $\mathit{58.83}$ & $\mathit{47.50}$ & $\mathit{31.74}$ & $\mathit{6.88}$ \\
\rowcolor{lightgray!30}\textbf{FunFace (ours)} & \multirow{-2}{*}{MS1MV3}   &  $\mathbf{68.68}$ & $\mathbf{78.25}$  & $\mathbf{35.58}$ & $\mathbf{47.16}$ & $\mathbf{59.69}$ & $\mathbf{47.98}$ & $\mathbf{32.70}$ & $\mathbf{8.85}$\\

\cmidrule(lr){1-10} 

Adaface\cite{adaface} &  & $\mathit{72.91}$ & $\mathit{81.87}$  & $\mathit{41.62}$ & $\mathit{53.67}$& $\mathit{67.93}$ & $\mathit{57.04}$ & $\mathit{41.20}$ & $\mathit{10.37}$ \\

\rowcolor{lightgray!30}\textbf{FunFace (ours)} & \multirow{-2}{*}{WebFace4M}   & $\mathbf{75.06}$ & $\mathbf{83.51}$  & $\mathbf{44.10}$ & $\mathbf{55.93}$& $\mathbf{71.12}$ & $\mathbf{61.48}$ & $\mathbf{46.41}$ & $\mathbf{16.18}$  \\

\bottomrule  
\end{NiceTabular}
}
\label{tab:dronesurf_and_survface}
\end{table*}

\section{Experiments \& Results}\label{sec:experiments_and_results}

\subsection{Experimental Setup}\label{sec:experiments_and_results:subsec:experimental_setup}

\noindent\textbf{Datasets.}  We divide the benchmarks according to their scale and difficulty into:$(i)$ small-scale, low difficulty LFW~\cite{lfw}, CFP-FP~\cite{cfp-fp}, CPLFW~\cite{cplfw}, AgeDB~\cite{agedb}, and CALFW~\cite{calfw} benchmarks, $(ii)$ large-scale, medium difficulty %
IJB-C~\cite{ijbc}, and $(iii)$ small-scale, high difficulty TinyFace~\cite{tinyface}, DroneSURF~\cite{dronesurf}, and SurvFace~\cite{survface}. More information about the benchmarks can be found in Table \ref{tab:dataset_info}. For training, we employed three commonly used face datasets: $(i)$ MS1MV2~\cite{ms1m, arcface}, $(ii)$ MS1MV3~\cite{ms1m, ms1mv3}, and $(iii)$ WebFace4M~\cite{webface260m}, containing $5.8$M, $5.1$M, and $4.2$M images of $85$k, $93$k and $205$k subjects, respectively. %

\noindent\textbf{Training Setting.} The training datasets have been preprocessed to a final size of $112\times112$ pixels using a standard five-point landmark alignment \cite{arcface} using the RetinaFace \cite{retinaface}, and MTCNN \cite{mtcnn} detectors. For the FR model, we make use of the ResNet-100 \cite{resnet} backbone and train it using SGD, with an initial learning rate of $1e^{-1}$, weight decay of $5e^{-4}$, and momentum of $0.9$. %
The learning rate is decreased by a factor of $10$ at the end of the $12$th, $20$th, and $24$th epoch. %
We use a mini-batch of $1024$ samples and train for $26$ epochs. 
For the implementation of FunFace, we closely follow the findings in \cite{adaface}, setting the hyperparameter $h$, which controls the spread of the normalized distributions to $0.333$, and the scaling hyperparameter $s$ to $64$.
The models were trained using a compute node %
equipped with two H100 80GB GPUs, an AMD EPYC 9124 64-core CPU, and 256GB of RAM.

\noindent\textbf{Augmentations.} In \cite{adaface} they show that while static margin loss functions do not see benefits from aggressive augmentations of training samples, the same does not necessarily hold for adaptive margins. Our focus is to excel in both high- and low-quality domains. Therefore, we extend the augmentations proposed in \cite{adaface}, i.e., cropping, rescaling, and photometric jittering, with noising, grayscaling, masking, and affine transformations. The use of augmentations allows the FR model to better adapt to a larger variety of image qualities, ensuring robust performance in low-quality domains.

Noising produces a weighted average of the original sample and a noise mask, sampled from a uniform or Gaussian distribution. Affine transformations can rotate, translate, and shear the sample, addressing potential issues of the face detection step. Grayscaling blends data from the RGB channels into a single grayscale channel. Masking obscures a rectangular area of the image and can imitate various occlusions due to face masks, glasses, hats, or any other potential objects blocking a part of the face. All operations are applied randomly with a probability of $20\%$, except for grayscaling, which is applied with a probability of $5\%$.

\begin{table*}[!t]
    \centering
    \caption{\textbf{Experimental results of the ablation study.} We present a study of the mixing parameter $\lambda$ and augmentations on the performance of the FunFace loss, across a wide range of benchmark datasets. We mark the \textbf{best}/\textit{second-best} result.}
      \renewcommand{\arraystretch}{1.05}
    \resizebox{0.95\linewidth}{!}{%
\begin{NiceTabular}{l  l | c | c | c | c | c || c | c | c || c }
\toprule
\multicolumn{2}{c}{\multirow{2}{*}{\textbf{Method}}} & \multirow{2}{*}{\textbf{LFW}} & \multirow{2}{*}{\textbf{CFP-FP}} & \multirow{2}{*}{\textbf{CPLFW}} & \multirow{2}{*}{\textbf{AgeDB}} & \multirow{2}{*}{\textbf{CALFW}} & \multicolumn{3}{c}{\textbf{IJB-C (TAR@FAR=?)}} & \textbf{TinyFace}  \\
& & & & & & & $1e^{-3}$ & $1e^{-4}$ & $1e^{-5}$ & Rank-1\\
\midrule
  \multicolumn{2}{c}{AdaFace}&  $99.85$& $99.23$& $94.63$& $\mathbf{97.93}$& $96.08$ & $\mathbf{97.36}$ &$95.37$ & $90.85$ &  $72.99$\\
  \cmidrule(lr){1-11}
  \multirow{5}{*}{\rotatebox[origin=c]{90}{FunFace($\lambda$)}}
  & \rowcolor{gray!20}$0.1$ & $\mathbf{99.85}$& $99.14$& $94.63$& $97.90$& $\mathbf{96.15}$ & $97.31$ & $\mathbf{95.62}$ & $90.81$ &  $\mathbf{73.31}$ \\
  & $0.3$ & $\mathit{99.85}$& $99.20$& $\mathit{94.77}$& $\mathit{97.92}$& $96.02$ & $97.28$ & $95.55$ & $\mathbf{91.81}$ & $\mathit{73.26}$\\
  & $0.5$ & $99.83$& $\mathbf{99.24}$& $\mathbf{94.78}$& $97.88$& $\mathit{96.10}$ & $\mathit{97.34}$ & $95.52$ & $90.33$ & $72.88$\\
  & $0.7$ & $99.85$& $99.21$& $94.63$& $97.83$& $96.12$ & $97.34$ & $\mathit{95.60}$ & $\mathit{91.23}$ & $72.83$\\
  & $0.9$ & $99.82$& $\mathit{99.23}$& $94.65$& $97.82$& $96.02$ & $97.31$ & $95.48$ & $91.23$ & $72.32$\\
  \midrule \midrule
  \multirow{2}{*}{\rotatebox[origin=c]{90}{/ Aug.}} & Ada. &  $\mathit{99.80}$& $\mathit{99.24}$& $\mathbf{94.62}$& $\mathbf{98.07}$& $\mathit{96.03}$ & $\mathit{97.08}$ & $\mathit{94.60}$ & $\mathit{87.07}$ &  $\mathit{71.03}$ \\
  & Fun. &  $\mathbf{99.85}$& $\mathbf{99.24}$& $\mathit{94.52}$& $\mathit{97.88}$& $\mathbf{96.03}$ & $\mathbf{97.23}$ & $\mathbf{95.36}$ & $\mathbf{89.88}$ & $\mathbf{71.14}$\\

\bottomrule  
\end{NiceTabular}
}\label{tab:mixing_parameter_ablation}

\end{table*}

\begin{table*}[!t]
    \centering
    \caption{\textbf{Comparison of the runtime between methods.} We present the results for three techniques, ArcFace, AdaFace, and FunFace, showcasing the differences in runtime measured in seconds.}
      \renewcommand{\arraystretch}{0.5}
    \resizebox{0.8\linewidth}{!}{%
\begin{NiceTabular}{l | r c l r c l r c l  }

\toprule

\textbf{Method} & \multicolumn{3}{c}{\textbf{ArcFace}\cite{adaface}} & \multicolumn{3}{c}{\textbf{AdaFace}\cite{adaface}} & \multicolumn{3}{c}{\rowcolor{lightgray!30}\textbf{FunFace}} \\
 \midrule

Time[s] ($\mu \pm \sigma$) & $0.0032$ & $\pm$ & $0.0021$ & $0.0075$ & $\pm$ & $0.0021$& \cellcolor{lightgray!30}$0.1220$ & \cellcolor{lightgray!30} $\pm$ & \cellcolor{lightgray!30} $0.0058$ \\
\bottomrule

\end{NiceTabular}
}
\label{tab:time_complexity}
\end{table*}

\subsection{Comparison with State-of-the-art}\label{sec:experiments_and_results:subsec:comparison_with_state_of_the_art}

In this section, we compare the results of the proposed FUNFace loss function with $13$ state-of-the-art approaches, including a variety of static and adaptive margin loss functions. We divide the experiments according to the difficulty of the benchmark datasets, as shown in Table \ref{tab:dataset_info}. %

\noindent\textbf{Low and Medium Difficulty.} Table \ref{tab:sota_small_and_mixed_scale} presents the experimental results across the low difficulty, i.e., LFW, CFP-FP, CPLFW, AgeDB, and CALFW, as well as the medium difficulty IJB-C benchmark. For the low difficulty benchmarks, we show the verification accuracy, while for the medium difficulty benchmark, we show the True Acceptance Rate (TAR) at three common False Acceptance Rates: $1e^{-4}$, $1e^{-5}$, and $1e^{-6}$. The methods are further divided according to the training dataset. From the results, we can see that the performance of FunFace is similar to current state-of-the-art methods such as AdaFace. While variations in specific scenarios can be seen, such as the increase in performance for the MS1MV3 dataset on all benchmarks, or a slight decrease on IJB-C, with WebFace4M, overall, the method performs well on both low and medium difficulty benchmarks.

\noindent\textbf{High Difficulty.} The experimental results for the low-quality (high difficulty) benchmarks, i.e., TinyFace, DroneSURF, and SurvFace, are shown in Tables \ref{tab:sota_low_quality} and \ref{tab:dronesurf_and_survface}. In Table \ref{tab:sota_low_quality}, we show the results on the TinyFace benchmark, using the Rank-$1$ and Rank-$5$ metrics. The results of competing methods are taken from \cite{adaface}. From the experimental results, we see that our method outperforms AdaFace across all training datasets, achieving state-of-the-art performance. This result showcases the power of biometric utility as a descriptor of sample quality and its importance in the training process of FR models. 
In Table \ref{tab:dronesurf_and_survface}, we show the results on both the DroneSURF and SurvFace benchmarks. Since results have not been reported on these benchmarks for any method, we evaluate them ourselves. We use models directly provided by the authors and closely follow the provided image preprocessing for each included method. We include results from three excellent adaptive loss functions, i.e., MagFace, ElasticFace, and AdaFace, and compare them to results from FunFace. For the DroneSURF benchmark, we divide the results into the Active scenario (DroneSURF-A) and the Passive scenario (DroneSURF-P), aligning with the division of data in the dataset. For both scenarios, we evaluate each frame individually, reporting the Rank-$1$ and Rank-$5$ metrics. For the SurvFace benchmark, we perform verification tests and report them using the True Acceptance Rate (TAR) at four False Acceptance Rates (FARs), i.e., $30\%$, $20\%$, $10\%$, and $1\%$, similar to the author suggested values. Unlike the authors of the benchmark, we do not, however, perform fine-tuning on the included training set of images. The results showcase a similar story as before. FunFace outperforms the included adaptive methods across all training datasets, achieving state-of-the-art performance. %

\subsection{Ablation Study}\label{sec:experiments_and_results:subsec:ablation_study}

In this section, we present an ablation study of the proposed FunFace loss functions, and two important factors surrounding it: $(i)$ the $\lambda$ parameter and $(ii)$ the image augmentations, as well as a comparison of the time complexity.

Experimental results across a wide-range of benchmarks are presented in Table \ref{tab:mixing_parameter_ablation}. We include results of a FR model trained using the AdaFace loss function, with our extended augmentations as well as using no augmentations, for better comparison with our proposed method. Focusing first on the effects of the mixing factor $\lambda$ on the performance of the model. On the low and medium difficulty benchmarks, the results seem to be inconclusive, with performance varying between individual benchmarks. On the high difficulty TinyFace benchmark, however, we can see a clear pattern between the value of $\lambda$ and the achieved performance of the FR model. We see that the inclusion of the certainty ratio improves performance over the AdaFace model, when using $\lambda=[0.1, 0.3]$. However, we can also see that the performance quickly drops with an increase of $\lambda$, showcasing the importance of the balance of information present in the feature norms and biometric utility of samples. Comparing the results of models trained with no augmentations using both AdaFace and FunFace, we see a similar result as when training using augmentations. The performance on low and medium-difficulty benchmarks closely matches between the adaptive functions, with a surprising result on IJB-C, where FunFace achieves a marginal improvement over AdaFace. For high difficulty images, we again see that FunFace outperforms AdaFace, but the difference is not as convincing as with the augmentations, showing the importance of lower quality samples during training of the model.

In Table \ref{tab:time_complexity} we show a comparison of the runtime between three methods: static margin ArcFace, adaptive AdaFace, and our proposed FunFace loss function. The presented times are measured across a single epoch over the WebFace4M dataset, using a batch size of $1024$. From the results, we can observe a drastic increase in the runtime of FunFace stemming from the costly computation of the closest negative neighbour of each sample in the batch, as well as the additional time to compute the normalized certainty ratio. Comparing AdaFace and ArcFace, the runtime is much closer, while, expectedly, ArcFace is still faster than AdaFace.

\section{Conclusion}\label{conclusion}

In this paper, we presented a new loss function, FunFace, focused on performance improvements in low-quality domains. %
FunFace builds on the success of AdaFace, which encodes information about the visual quality into the margin using feature norms. We show that the biometric utility is a superior measure of sample difficulty for recognition compared to visual quality estimators, and that it cannot be accurately portrayed using exisiting information present in the feature norms. This motivates us to include an additional source of information directly related to the biometric utility into the FunFace margin function. We employ the CR-FIQA technique to estimate the utility of samples online during training and combine it with the feature norm information into the final margin value. %
From the experimental evaluation of the method, we can observe that the performance for high and mixed quality images is comparable to state-of-the-art techniques such as AdaFace. For images of lower quality, as showcased by high difficulty benchmarks, we see a dramatic improvement in quality, stemming from the addition of the extended augmentations and the predictive power of the certainty ratio for the biometric utility of samples.

\noindent\textbf{Limitations \& Future Work.} This work focuses on improvements in low-quality domains, as such, its performance in high- and medium-quality domains does not show a clear improvement over state-of-the-art methods. Another clear downside of the approach is its heavier computational runtime, which significantly affects the training time of models. Additionally, the provided analysis shows a benefit of using extended augmentations of input samples, but explores the correlation only briefly. The proposed loss function is derivative work, extending on previous work, namely AdaFace and CR-FIQA, and therefore does not explore all hyperparameters and variables discussed already in the mentioned works.

\section{Ethical Statement}\label{ethical_statement}

\noindent\textbf{A. Ethical Review Boards} The present work develops a face-recognition training objective (FunFace) and evaluates it on established, publicly available training and benchmark datasets. We did not recruit participants, perform interventions, or conduct user studies, and we did not collect new biometric data; therefore, formal oversight by an ethical review board was not required for the work reported here.%

\noindent\textbf{B. Potential Harms to Human Subjects} Although no new data were collected, face datasets inherently contain biometric information, and misuse or leakage could impact individuals’ privacy. In line with the conference guidance on informed consent and the protection of privacy and confidentiality, we do not release any raw face imagery or identity annotations beyond what is already provided by the original dataset distributors, and we handle data according to the corresponding dataset terms.

\noindent\textbf{C. Potential Negative Societal Impacts} Our primary goal is to improve face recognition robustness in challenging, low-quality scenarios. Such improvements can be beneficial in benign applications (e.g., robust authentication in unconstrained conditions), but they may also enhance surveillance capabilities, which can be used in ways that limit rights, privacy, or exacerbate discrimination. We explicitly acknowledge this dual-use risk and do not endorse deployments that violate applicable laws, human rights, or privacy expectations.

\noindent\textbf{D. Risk-Mitigation Strategies} To mitigate risks of misuse and misunderstanding, we (i) clearly scope our contribution as a training objective and report limitations, (ii) commit to releasing code and trained models to support transparency and reproducibility, and (iii) refrain from distributing withdrawn or restricted datasets. In particular, to enable fair comparison with prior work, we report results using MS-Celeb-derived training data, noting that MS-Celeb has been withdrawn; we do not redistribute this data and encourage the use of currently available alternatives when possible. These steps align with the emphasis on risk mitigation via controlled access to research materials and clear communication of intended use and limitations.

{\small
\bibliographystyle{ieee}
\bibliography{egbib}

\begin{thebibliography}{10}\itemsep=-1pt

\bibitem{partialfc}
X.~An, X.~Zhu, Y.~Gao, Y.~Xiao, Y.~Zhao, Z.~Feng, L.~Wu, B.~Qin, M.~Zhang, D.~Zhang, et~al.
\newblock {Partial FC: Training 10 Million Identities on a Single Machine}.
\newblock In {\em Proceedings of the CVF/IEEE International Conference on Computer Vision (ICCV)}, pages 1445--1449, 2021.

\bibitem{froq}
{\v{Z}}.~Babnik, D.~K. Jain, P.~Peer, and V.~{\v{S}}truc.
\newblock {FROQ: Observing Face Recognition Models for Efficient Quality Assessment}.
\newblock {\em arXiv preprint arXiv:2509.17689}, 2025.

\bibitem{faceqan}
{\v{Z}}.~Babnik, P.~Peer, and V.~{\v{S}}truc.
\newblock {FaceQAN: Face Image Quality Assessment through Adversarial Noise Exploration}.
\newblock In {\em Proceedings of the IAPR International Conference on Pattern Recognition (ICPR)}, pages 748--754, 2022.

\bibitem{ediffiqa}
{\v{Z}}.~Babnik, P.~Peer, and V.~{\v{S}}truc.
\newblock {eDifFIQA: Towards Efficient Face Image Quality Assessment Based on Denoising Diffusion Probabilistic Models}.
\newblock {\em IEEE Transactions on Biometrics, Behavior, and Identity Science (TBIOM)}, 6(4):458--474, 2024.

\bibitem{elasticface}
F.~Boutros, N.~Damer, F.~Kirchbuchner, and A.~Kuijper.
\newblock {ElasticFace: Elastic Margin Loss for Deep Face Recognition}.
\newblock In {\em Proceedings of the CVF/IEEE International Conference on Computer Vision and Pattern Recognition (CVPR) Workshops}, 2022.

\bibitem{cr-fiqa}
F.~Boutros, M.~Fang, M.~Klemt, B.~Fu, and N.~Damer.
\newblock {CR-FIQA: Face Image Quality Assessment by Learning Sample Relative Classifiability}.
\newblock In {\em Proceedings of the CVF/IEEE International Conference on Computer Vision and Pattern Recognition (CVPR)}, 2023.

\bibitem{tinyface}
Z.~Cheng, X.~Zhu, and S.~Gong.
\newblock {Low-Resolution Face Recognition}.
\newblock In {\em Proceedings of the Asian Conference on Computer Vision (ACCV)}, pages 605--621. Springer, 2018.

\bibitem{survface}
Z.~Cheng, X.~Zhu, and S.~Gong.
\newblock {Surveillance face recognition challenge}.
\newblock {\em arXiv preprint arXiv:1804.09691}, 2018.

\bibitem{briar}
D.~Cornett, J.~Brogan, N.~Barber, D.~Aykac, S.~Baird, N.~Burchfield, C.~Dukes, A.~Duncan, R.~Ferrell, J.~Goddard, et~al.
\newblock {Expanding Accurate Person Recognition to New Altitudes and Ranges: The Briar Dataset}.
\newblock In {\em Proceedings of the CVF/IEEE Winter Conference on Applications of Computer Vision (WACV)}, pages 593--602, 2023.

\bibitem{retinaface}
J.~Deng, J.~Guo, E.~Ververas, I.~Kotsia, and S.~Zafeiriou.
\newblock {RetinaFace: Single-Shot Multi-Level Face Localisation in the Wild}.
\newblock In {\em Proceedings of the CVF/IEEE International Conference on Computer Vision and Pattern Recognition (CVPR)}, pages 5203--5212, 2020.

\bibitem{arcface}
J.~Deng, J.~Guo, N.~Xue, and S.~Zafeiriou.
\newblock {Arcface: Additive Angular Margin Loss for Deep Face Recognition}.
\newblock In {\em Proceedings of the CVF/IEEE International Conference on Computer Vision and Pattern Recognition (CVPR)}, pages 4690--4699, 2019.

\bibitem{vplarcface}
J.~Deng, J.~Guo, J.~Yang, A.~Lattas, and S.~Zafeiriou.
\newblock {Variational Prototype Learning for Deep Face Recognition}.
\newblock In {\em Proceedings of the CVF/IEEE Conference on Computer Vision and Pattern Recognition (CVPR)}, pages 11906--11915, 2021.

\bibitem{ms1mv3}
J.~Deng, J.~Guo, D.~Zhang, Y.~Deng, X.~Lu, and S.~Shi.
\newblock {Lightweight Face Recognition Challenge}.
\newblock In {\em {Proceedings of the CVF/IEEE International Conference on Computer Vision Workshops (ICCVW)}}, pages 0--0, 2019.

\bibitem{fr_survey}
H.~Du, H.~Shi, D.~Zeng, X.-P. Zhang, and T.~Mei.
\newblock {The Elements of End-to-End Deep Face Recognition: A Survey of Recent Advances}.
\newblock {\em ACM computing surveys (CSUR)}, 54(10s):1--42, 2022.

\bibitem{ms1m}
Y.~Guo, L.~Zhang, Y.~Hu, X.~He, and J.~Gao.
\newblock {MS-Celeb-1M: A Dataset and Benchmark for Large-Scale Face Recognition}.
\newblock In {\em Proceedings of the European Conference on Computer Vision (ECCV)}, pages 87--102. Springer, 2016.

\bibitem{resnet}
K.~He, X.~Zhang, S.~Ren, and J.~Sun.
\newblock {Deep Residual Learning for Image Recognition}.
\newblock In {\em Proceedings of the CVF/IEEE Conference on Computer Vision and Pattern Recognition}, pages 770--778, 2016.

\bibitem{faceqnet2}
J.~Hernandez-Ortega, J.~Galbally, J.~Fierrez, R.~Haraksim, and L.~Beslay.
\newblock {FaceQnet: Quality Assessment for Face Recognition Based on Deep Learning}.
\newblock In {\em Proceedings of the IAPR International Conference on Biometrics (ICB)}, pages 1--8, 2019.

\bibitem{lfw}
G.~B. Huang, M.~Ramesh, T.~Berg, and E.~Learned-Miller.
\newblock {Labeled Faces in the Wild: A Database for Studying Face Recognition in Unconstrained Environments}.
\newblock Technical Report 07-49, University of Massachusetts, Amherst, October 2007.

\bibitem{curricularface}
Y.~Huang, Y.~Wang, Y.~Tai, X.~Liu, P.~Shen, S.~Li, J.~Li, and F.~Huang.
\newblock {CurricularFace: Adaptive Curriculum Learning Loss for Deep Face Recognition}.
\newblock In {\em Proceedings of the CVF/IEEE International Conference on Computer Vision and Pattern Recognition (CVPR)}, pages 5901--5910, 2020.

\bibitem{eval2}
{ISO/IEC DIS 29794-1, Biometric Sample Quality}.
\newblock Standard, International Organization for Standardization (ISO), 2022.

\bibitem{dronesurf}
I.~Kalra, M.~Singh, S.~Nagpal, R.~Singh, M.~Vatsa, and P.~Sujit.
\newblock {DroneSURF: Benchmark Dataset for Drone-Based Face Recognition}.
\newblock In {\em Proceedings of the IEEE International Conference on Automatic Face \& Gesture Recognition (FG)}, pages 1--7. IEEE, 2019.

\bibitem{adaface}
M.~Kim, A.~K. Jain, and X.~Liu.
\newblock {AdaFace: Quality Adaptive Margin for Face Recognition}.
\newblock In {\em Proceedings of the CVF/IEEE International Conference on Computer Vision and Pattern Recognition (CVPR)}, pages 18750--18759, 2022.

\bibitem{broadface}
Y.~Kim, W.~Park, and J.~Shin.
\newblock {BroadFace: Looking at Tens of Thousands of People at Once for Face Recognition}.
\newblock In {\em Proceedings of the European Conference on Computer Vision (ECCV)}, pages 536--552. Springer, 2020.

\bibitem{xqlfw}
M.~Knoche, S.~Hormann, and G.~Rigoll.
\newblock {Cross-Quality LFW: A Database for Analyzing Cross-Resolution Image Face Recognition in Unconstrained Environments}.
\newblock In {\em Proceedings of the IEEE International Conference on Automatic Face and Gesture Recognition (FG)}, pages 1--5, 2021.

\bibitem{sphereface}
W.~Liu, Y.~Wen, Z.~Yu, M.~Li, B.~Raj, and L.~Song.
\newblock {Sphereface: Deep Hypersphere Embedding for Face Recognition}.
\newblock In {\em Proceedings of the CVF/IEEE Conference on Computer Vision and Pattern Recognition (CVPR)}, pages 212--220, 2017.

\bibitem{rankiqa}
X.~Liu, J.~Van De~Weijer, and A.~D. Bagdanov.
\newblock {RankIQA: Learning from Rankings for No-Reference Image Quality Assessment}.
\newblock In {\em Proceedings of the CVF/IEEE International Conference on Computer Vision (ICCV)}, pages 1040--1049, 2017.

\bibitem{swintransformer}
Z.~Liu, Y.~Lin, Y.~Cao, H.~Hu, Y.~Wei, Z.~Zhang, S.~Lin, and B.~Guo.
\newblock {SWIN Transformer: Hierarchical Vision Transformer Using Shifted Windows}.
\newblock In {\em Proceedings of the CVF/IEEE International Conference on Computer Vision}, pages 10012--10022, 2021.

\bibitem{ijbc}
B.~Maze, J.~Adams, J.~A. Duncan, N.~Kalka, T.~Miller, C.~Otto, A.~K. Jain, W.~T. Niggel, J.~Anderson, J.~Cheney, et~al.
\newblock {IARPA Janus Benchmark-C: Face Dataset and Protocol}.
\newblock In {\em Proceedings of the International Conference on Biometrics (ICB)}, pages 158--165, 2018.

\bibitem{magface}
Q.~Meng, S.~Zhao, Z.~Huang, and F.~Zhou.
\newblock {MagFace: A Universal Representation for Face Recognition and Quality Assessment}.
\newblock In {\em Proceedings of the CVF/IEEE International Conference on Computer Vision and Pattern Recognition (CVPR)}, pages 14225--14234, 2021.

\bibitem{brisque}
A.~Mittal, A.~K. Moorthy, and A.~C. Bovik.
\newblock {No-Reference Image Quality Assessment in the Spatial Domain}.
\newblock {\em IEEE Transactions on Image Processing (TIP)}, 21(12):4695--4708, 2012.

\bibitem{niqe}
A.~Mittal, R.~Soundararajan, and A.~C. Bovik.
\newblock {Making a “Completely Blind” Image Quality Analyzer}.
\newblock {\em IEEE Signal Processing Letters (SPL)}, 20(3):209--212, 2012.

\bibitem{agedb}
S.~Moschoglou, A.~Papaioannou, C.~Sagonas, J.~Deng, I.~Kotsia, and S.~Zafeiriou.
\newblock {AgeDB: the First Manually Collected, in-the-Wild Age Database}.
\newblock In {\em {Proceedings of the CVF/IEEE Conference on Computer Vision and Pattern Recognition Workshops (CVPRW)}}, pages 51--59, 2017.

\bibitem{petalface}
K.~Narayan, N.~G. Nair, J.~Xu, R.~Chellappa, and V.~M. Patel.
\newblock {PetalFace: Parameter Efficient Transfer Learning for Low-Resolution Face Recognition}.
\newblock In {\em Proceedings of the CVF/IEEE Winter Conference on Applications of Computer Vision (WACV)}, pages 804--814. IEEE, 2025.

\bibitem{sdd-fiqa}
F.-Z. Ou, X.~Chen, R.~Zhang, Y.~Huang, S.~Li, J.~Li, Y.~Li, L.~Cao, and Y.-G. Wang.
\newblock {SDD-FIQA: Unsupervised Face Image Quality Assessment with Similarity Distribution Distance}.
\newblock In {\em Proceedings of the CVF/IEEE International Conference on Computer Vision and Pattern Recognition (CVPR)}, pages 7670--7679, 2021.

\bibitem{clibfiqa}
F.-Z. Ou, C.~Li, S.~Wang, and S.~Kwong.
\newblock {CLIB-FIQA: Face Image Quality Assessment with Confidence Calibration}.
\newblock In {\em Proceedings of the CVF/IEEE Conference on Computer Vision and Pattern Recognition (CVPR)}, pages 1694--1704, 2024.

\bibitem{survey}
T.~Schlett, C.~Rathgeb, O.~Henniger, J.~Galbally, J.~Fierrez, and C.~Busch.
\newblock {Face Image Quality Assessment: A Literature Survey}.
\newblock {\em ACM Computing Surveys (CSUR)}, 54(10s):1--49, 2022.

\bibitem{eval1}
T.~Schlett, C.~Rathgeb, J.~Tapia, and C.~Busch.
\newblock {Considerations on the Evaluation of Biometric Quality Assessment Algorithms}.
\newblock {\em arXiv preprint arXiv:2303.13294}, 2023.

\bibitem{facenet}
F.~Schroff, D.~Kalenichenko, and J.~Philbin.
\newblock {Facenet: A Unified Embedding for Face Recognition and Clustering}.
\newblock In {\em Proceedings of the CVF/IEEE Conference on Computer Vision and Pattern Recognition (CVPR)}, pages 815--823, 2015.

\bibitem{cfp-fp}
S.~Sengupta, J.~C. Cheng, C.~D. Castillo, V.~M. Patel, R.~Chellappa, and D.~W. Jacobs.
\newblock {Frontal to Profile Face Verification in the Wild}.
\newblock In {\em Proceedings of the IEEE Winter Conference on Applications of Computer Vision (WACV)}, 2016.

\bibitem{url}
Y.~Shi, X.~Yu, K.~Sohn, M.~Chandraker, and A.~K. Jain.
\newblock {Towards Universal Representation Learning for Deep Face Recognition}.
\newblock In {\em Proceedings of the CVF/IEEE Conference on Computer Vision and Pattern Recognition (CVPR)}, pages 6817--6826, 2020.

\bibitem{serfiq}
P.~Terhorst, J.~N. Kolf, N.~Damer, F.~Kirchbuchner, and A.~Kuijper.
\newblock {SER-FIQ: Unsupervised Estimation of Face Image Quality Based on Stochastic Embedding Robustness}.
\newblock In {\em Proceedings of the CVF/IEEE International Conference on Computer Vision and Pattern Recognition (CVPR)}, pages 5651--5660, 2020.

\bibitem{transformer}
A.~Vaswani, N.~Shazeer, N.~Parmar, J.~Uszkoreit, L.~Jones, A.~N. Gomez, {\L}.~Kaiser, and I.~Polosukhin.
\newblock {Attention is All You Need}.
\newblock {\em Advances in Neural Information Processing Systems (NeurIPS)}, 30, 2017.

\bibitem{cosface}
H.~Wang, Y.~Wang, Z.~Zhou, X.~Ji, D.~Gong, J.~Zhou, Z.~Li, and W.~Liu.
\newblock {CosFace: Large Margin Cosine Loss for Deep Face Recognition}.
\newblock In {\em Proceedings of the CVF/IEEE International Conference on Computer Vision and Pattern Recognition (CVPR)}, pages 5265--5274, 2018.

\bibitem{mvsoftmax}
X.~Wang, S.~Zhang, S.~Wang, T.~Fu, H.~Shi, and T.~Mei.
\newblock {Mis-classified Vector Guided Softmax Loss for Face Recognition}.
\newblock In {\em Proceedings of the AAAI Conference on Artificial Intelligence}, volume~34, pages 12241--12248, 2020.

\bibitem{mtcnn}
J.~Xiang and G.~Zhu.
\newblock {Joint Face Detection and Facial Expression Recognition with MTCNN}.
\newblock In {\em 2017 4th International Conference on Information Science and Control Engineering (ICISCE)}, pages 424--427. IEEE, 2017.

\bibitem{cplfw}
T.~Zheng and W.~Deng.
\newblock {Cross-Pose LFW: A Database for Studying Cross-Pose Face Recognition in Unconstrained Environments}.
\newblock Technical Report 18-01, Beijing University of Posts and Telecommunications, February 2018.

\bibitem{calfw}
T.~Zheng, W.~Deng, and J.~Hu.
\newblock {Cross-Age {LFW:} {A} Database for Studying Cross-Age Face Recognition in Unconstrained Environments}.
\newblock {\em CoRR}, abs/1708.08197, 2017.

\bibitem{webface260m}
Z.~Zhu, G.~Huang, J.~Deng, Y.~Ye, J.~Huang, X.~Chen, J.~Zhu, T.~Yang, J.~Lu, D.~Du, et~al.
\newblock {WebFace260M: A Benchmark Unveiling the Power of Million-Scale Deep Face Recognition}.
\newblock In {\em Proceedings of the CVF/IEEE Conference on Computer Vision and Pattern Recognition (CVPR)}, pages 10492--10502, 2021.

\end{thebibliography}
}

\end{document}